\documentclass{article}

\usepackage{PRIMEarxiv}

\usepackage[utf8]{inputenc} % allow utf-8 input
\usepackage[T1]{fontenc}    % use 8-bit T1 fonts
\usepackage{hyperref}       % hyperlinks
\usepackage{url}            % simple URL typesetting
\usepackage{booktabs}       % professional-quality tables
\usepackage{amsfonts}       % blackboard math symbols
\usepackage{nicefrac}       % compact symbols for 1/2, etc.
\usepackage{microtype}      % microtypography
\usepackage{lipsum}
\usepackage{tabularx}
\usepackage{bm}
\usepackage{amsmath}
\usepackage{fancyhdr}       % header
\usepackage{graphicx}       % graphics
\graphicspath{{media/}}     % organize your images and other figures under media/ folder
\usepackage[export]{adjustbox}

\title{On Training of Kolmogorov-Arnold Networks
}

\author{
  Shairoz Sohail \\
  University of Illinois, Urbana-Champaign \\
  Urbana-Champaign, IL\\
  \texttt{shairozsohail@gmail.com} \\
}

\begin{document}
\maketitle

\begin{abstract}
Kolmogorov-Arnold Networks have recently been introduced as a flexible alternative to multi-layer Perceptron architectures. In this paper, we examine the training dynamics of different KAN architectures and compare them with corresponding MLP formulations. We train with a variety of different initialization schemes, optimizers, and learning rates, as well as utilize back propagation free approaches like the HSIC Bottleneck. We find that (when judged by test accuracy) KANs are an effective alternative to MLP architectures on high-dimensional datasets and have somewhat better parameter efficiency, but suffer from more unstable training dynamics. Finally, we provide recommendations for improving training stability of larger KAN models. 
%\newline
%Code available at github.com
\end{abstract}

% keywords can be removed

\section{Introduction}
For the last decade, the Perceptron \cite{Perceptron} has served as the defacto building block of deep neural networks that utilize a fully connected layer in their architecture. Recent breakthroughs in large language models are due partly to large stacks of Transformer \cite{attention} units, which also utilize Perceptrons as part of their internal machinery. The popularity of Perceptron based architectures can be attributed mainly to two things: flexibility for learning non-linear functions \cite{approx}, and an inherent ability to parallelize on modern GPU architecture \cite{parallel}. While Perceptron-free architectures have been shown to display impressive performance \cite{fully_conv}, the fully-connected Perceptron layer remains a staple in high performing models.

Recently, a viable alternative to the Perceptron unit for fully connected layers has been proposed - the Kolmogorov-Arnold unit \cite{kan}. This unit relies on the Kolmogorov-Arnold representation theorem \cite{kan-theorem} to fit a B-spline \cite{bspline} to input data, with the claim that this achieves performance on par or above that of the Perceptron when used in fully connected layers \cite{kan}. However, all recent work on KANs has relied on training schemes optimized over years of training Perceptron based networks, which may be sub-optimal to yield the best performance out of KAN architectures. 

Our contributions are as follows:

\begin{itemize}

\item We test several combinations of initialization, optimizer, and learning rate on KANs of different sizes using a variety of real world datasets, then compare with MLPs (fixing the number of layers or the number of parameters). 

\item We attempt to train KANs utilizing the HSIC Bottleneck \cite{hsic} (an alternative to back propagation).

\item We explore KAN scalability with regards to intrinsic dimension, width, and depth.

\item We explore additional metrics of KANs compared to MLP architectures including a novel, dataset-agnostic efficiency metric.

\end{itemize}

Our work can be seen as an extension of \cite{kan_suitability}, where experiments are conducted utilizing much larger models (10m+ parameters), additional datasets, and more training schemes. We restrict our attention to  fully connected architectures for ease of comparison. The focus of this paper is on discovering optimal training schemes for fully connected KANs while comparing with the performance of equivalently trained MLPs. 

\section{Background and Related Work}
\label{sec:headings}

\subsection{B-splines}
A \textbf{spline} \cite{splines} of degree k is a function defined piecewise by polynomials of degree $\leq k$. Spline functions are usually designed to smoothly interpolate complicated continuous functions while avoiding some of the issues (such as Runge's phenomona \cite{runge}) posed by fitting a single, high dimensional polynomial. In fact, it can be shown that any continuous function defined completely over a real-valued range $[a,b]$ admits such a decomposition into degree $n \geq 1$ polynomial functions, with uniform convergence as n goes to infinity \cite{kan-theorem}. 

For any specific spline of degree $k$, it can be represented as a linear combination of \textbf{B-splines} ("basis splines") of degree $k$. This means that for an unknown function $f(x)$, a degree $k$ spline $S(x)$ and sequence of degree $k$ B-splines $\{B_1(x)... B_n(x)\}$, we have:

\begin{equation} \label{eq1}
\begin{split}
f(x) \approx S(x) = \sum_{i = 1}^{n}c_i B_i(x)
\end{split}
\end{equation}

for $x \in [a,b]$. Thus, B-splines are a powerful starting point when one wishes to fit a complex and partially observed function. 

\subsection{The Kolmogorov-Arnold unit}

A single Kolmogorov-Arnold unit is defined using the following function:

\begin{equation} \label{eq2}
\begin{split}
 {\phi(x)  = {w_b}{\alpha(x)} + w_s}{\sum_{i} c_i B_i(x)}
\end{split}
\end{equation}

where $w_b, w_s$, and $c_i$ are learnable parameters, and $\alpha(x)$ is a non-linear activation function (such as $silu(x) = \frac{x}{1+e^{-x}}$). Using the notation from (1), we have:

\begin{equation} \label{eq2}
\begin{split}
\phi(x)  = {w_b}{\alpha(x) + w_s}{S(x)}
\end{split}
\end{equation}

Similar to MLPs, Kolmogorov-Arnold units can be placed in "layers" and composed. If we let $\phi_{i}^{l}$ represent the $i^{th}$ KA unit in layer $l$, then:

\begin{equation} \label{eq2}
\begin{split}
\phi_{i}^{l} = \sum_{i=1}^{n_{l-1}} \phi_{i,j}^{l-1}(x_{i}^{l-1})
\end{split}
\end{equation}

where $n_{l-1}$ is the number of KA units in the previous layer, and $x_{i}^{l-1}$ is the activation from the $i^{th}$ unit of the previous layer (or the $i^{th}$ input feature, when $l=1$). Thus, if we let $\bm{\Phi_l}$ denote a column vector $<\phi_{l}^1... \phi_{l}^{n_l}>$ of units in layer $l$, we can write a Kolmogorov-Arnold Network with $K$ layers as:

\begin{equation} \label{eq2}
\begin{split}
 KAN(x) = (\bm{\Phi_K} \circ \bm{\Phi_{K-1}} ...   \circ \bm{\Phi_1})(x)
\end{split}
\end{equation}

\newpage

\section{Method}
\label{sec:headings}
\subsubsection{Architecture Choices}
We test a shallow, medium, and deep architecture for each task (sizes are task dependent, see appendix 1.1 for details on width and depth). For each network, we keep the width and depth constant, but models can have more or less parameters depending on the parameterization of their units (a single KA-3 unit has more parameters than a single Perceptron unit). Note that the chosen widths are based on previous work with MLPs \cite{how_far_fcn}, and exploring large width KANs remains an open area of research. To help make MLP architectures more comparable to their KAN counterparts, a series of wide MLP architectures is also tested, where the width of each layer from the corresponding (small,medium,large) architecture is doubled but depth is kept the same. All networks utilize ReLU activation functions and a dropout probability of 0.2. For each network and architecture we vary the datasets and training schemes, as described below.

\subsubsection{Datasets}
We perform experiments using a variety of datasets spanning image classification, sentiment prediction, and scientific (tabular) data. We perform no image augmentation \cite{augmentation} or mixup \cite{mixup}. Below is a breakdown of the datasets:

\begin{table}[h]
\begin{tabular}{|l|l|l|l|l|}
\hline
Dataset       & Features & Training Examples & Testing Examples & Type                 \\ \hline
MNIST \cite{mnist}         & 28*28    & 60000             & 10000            & Image Classification \\ \hline
Fashion-MNIST \cite{fashion} & 28*28    & 60000             & 10000            & Image Classification \\ \hline
CIFAR-10 \cite{cifar}     & 3*28*28  & 50000             & 10000            & Image Classification \\ \hline
IMDB \cite{imdb}         & 5000     & 25000             & 25000            & Sentiment Prediction \\ \hline
HIGGS  \cite{higgs}       & 28       & 50000             & 50000            & Scientific Data      \\ \hline
\end{tabular}
\end{table}

\subsection{Training Schemes}
We define a training scheme for a particular model and dataset as the set of \{initialization, optimizer, initial learning rate, batch size, stopping criteria\} with the belief that this defines the training dynamics for a given model architecture and dataset. There is evidence to support this \cite{activations}\cite{optimizers}, as well as significant research performed on the effect of each of these individual factors on overall training success for MLPs \cite{training_factors}. We do not modify default learning rate schedules (which have shown to also have an effect on performance \cite{lr_schedule}).

We vary training schemes by testing three initialization schemes, three optimizers, and three learning rates, while keeping the batch size and stopping criteria fixed. This leads to 27 runs of each model on each dataset. We then perform experiments with the HSIC Bottleneck \cite{hsic}, a backpropagation-free algorithm. Finally, we take the best training schemes from these runs and perform additional experiments, varying the intrinsic dimension and activation function of the KANs to attempt further improvement. 

\begin{table}[h]
\begin{tabular}{ll}
Models:         \& \{MLP, KAN, MLP-Wide\}  \\
Initialization:         \& \{Kaiming-Normal, Kaiming-Uniform, Orthogonal\}  \\
Optimizers:     \& \{SGD, SGD-M, AdAM\}    \\
Activtions:     \& \{ReLU, Cosine\}        \\
Learning Rates: \& \{0.05, 0.005, 0.0005\}
\end{tabular}
\end{table}

\section{Results}
\subsubsection{Training with Backpropagation}
We observe that both MLPs and KANs achieve comparable accuracies across a variety of datasets when compared using fixed network architectures. However, KANs contain more parameters than MLPs even when network width and depth are fixed (because of the additional intrinsic parameters of KA-units). When MLPs are scaled up (by increasing layer width) to match the number of parameters of a KAN of the same depth, performance of the wide MLPs is comparable (HIGGS) or better (MNIST, Fashion MNIST, CIFAR10, IMDB) than KANs. Generally, MLP architectures outperform KAN architectures of the same parameter count as measured by accuracy and training time when back propagation is used (see barplots for "kan 2layers" vs. "mlpWide 2layers"). 

\newpage 
\begin{figure}[h]
\includegraphics[width=1\textwidth]{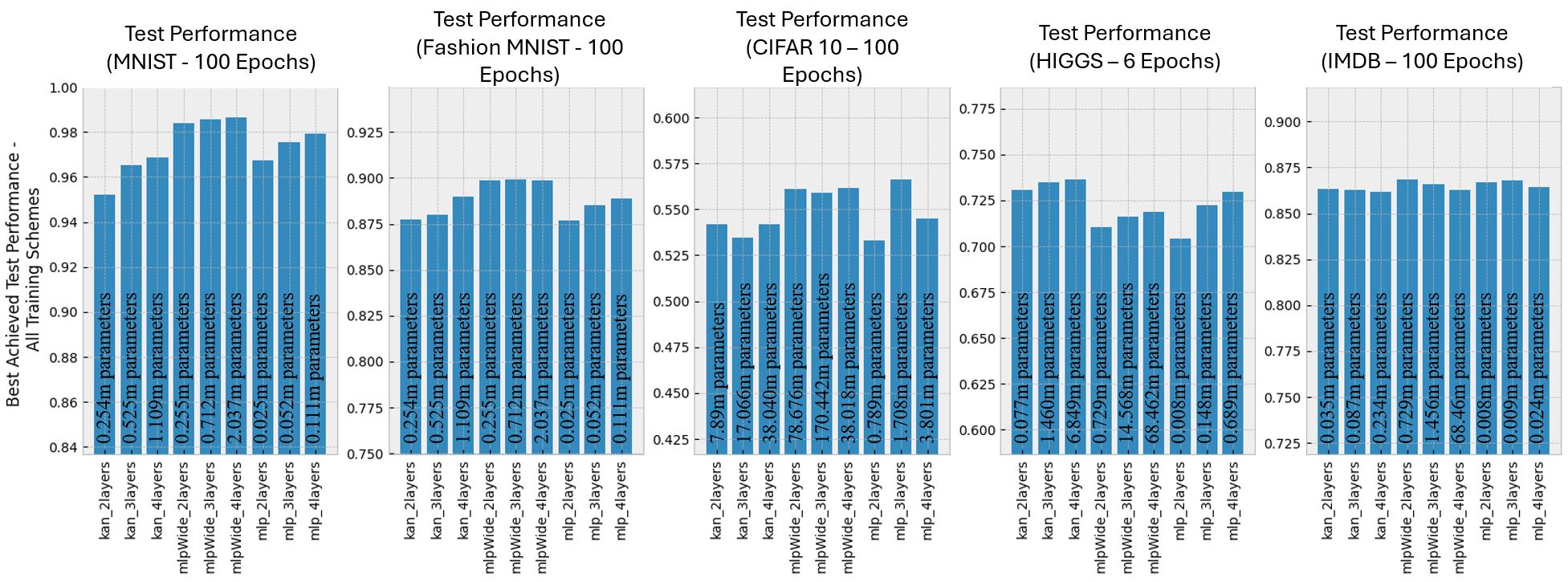}
\caption{Test accuracy of KANs and MLPs utilizing backpropagation}
\centering
\end{figure}

Additionally, KANs seem to have a higher tendency to overfit (as measured by the difference between training and test accuracy) than MLPs for all training schemes. This necessitates additional caution in training large KANs to prevent overfitting early in the training process. In general, large KANs seem to perform best when utilizing a Kaiming Normal \cite{kaiming_normal} weight initialization, an adaptive optimizer (such as AdAM), and low initial learning rate (<= 5e-4). Looking across datasets, another trend seems to emerge: final performance of KANs is more sensitive to choice of initialization, optimizer, and learning rate. For example, below is a boxplot of MNIST and Fashion MNIST performance under all backprop training schemes (varying initialization, optimzier, and learning rate):

\begin{figure}[h]
\centering
\includegraphics[scale=0.25]{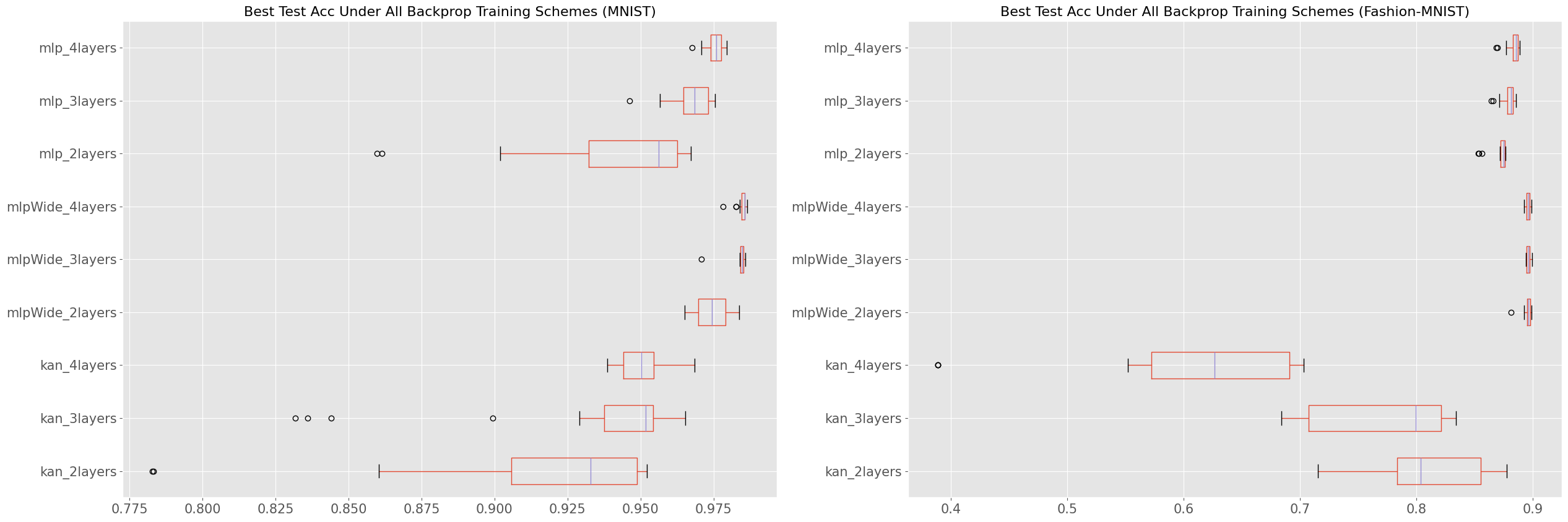}
\caption{Test accuracy of KANs and MLPs utilizing backpropagation}
\centering
\end{figure}

This indicates that it may be prudent to evaluate additional regularization and training protocols, including back-propagation free approaches. This may help in fully utilizing the expressive power of KAN layers.  

\subsubsection{Training with the HSIC Bottleneck}
One way to sidestep some of the challenges of training KANs with backpropagation is to use a backpropagation-free training algorithm. The HSIC Bottleneck \cite{hsic} utilizes a layer-by-layer joint information objective, and has been shown to successfully train networks with large parameter counts while avoiding some of the issues with back-propagation such as vanishing gradients, and leading to more stable and rapid  convergence. We hypothesize that given their tendency to overfit and sensitive dependence on optimizer, the HSIC Bottleneck may provide more stable and consistent training of KANs. We repeat the experiments above, except utilizing the HSIC objective instead of standard backpropagation.

\newpage

\begin{figure}[h]
\centering
\includegraphics[width=1\textwidth]{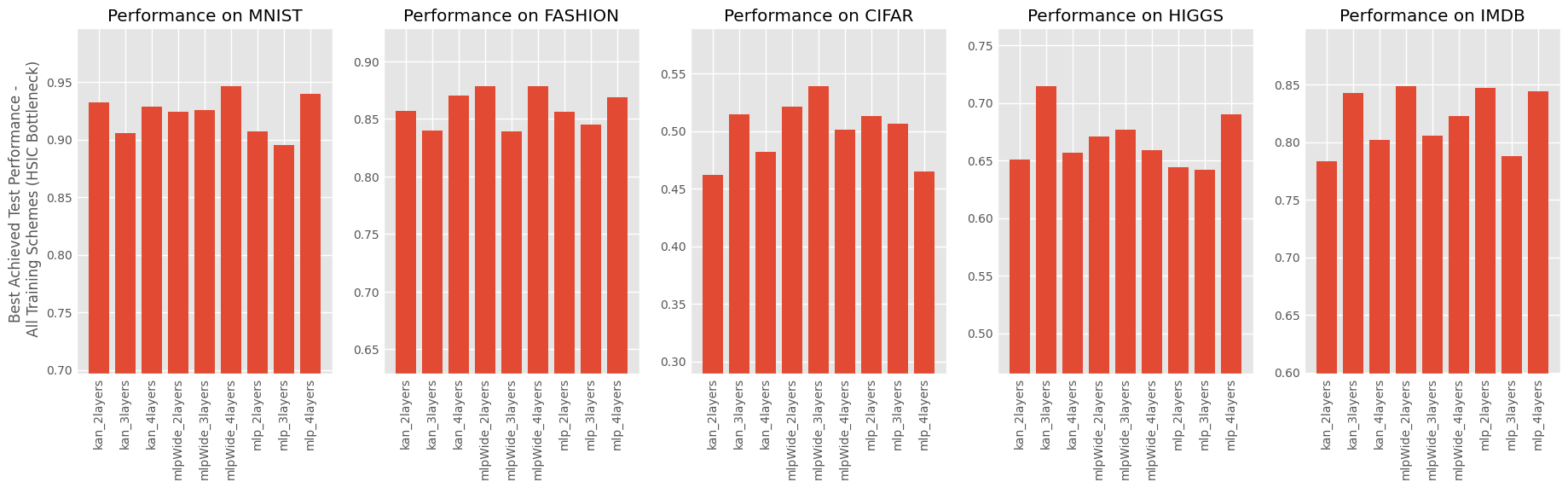}
\caption{Test accuracy of KANs and MLPs utilizing the HSIC Bottleneck}
\centering
\end{figure}

Unfortunately, performance trends of KANs compared to MLPs remain similar when HSIC Bottleneck training is used, with marginally worse final accuracies. 

\section{Additional Experiments}

\subsection{The Role of Activation Functions}
All initial training runs were performed with GELU activation functions, as in the original KAN architecture \cite{kan}. For this section we extract the best training scheme for each dataset's KAN and vary the activation function. Besides the GELU, we also evaluate the ELU and SILU activation functions, which (similar to GELU) attempt to mitigate the "dead neuron" \cite{deadneurons} effect seen in ReLU activated networks, where certain neurons become stuck at producing zero valued outputs. While dropout and batch normalization can help with this \cite{deadneuron_bn}, varying activation functions usually provides a synergistic effect. 

\begin{figure}[h]
\centering
\includegraphics[scale=0.30]{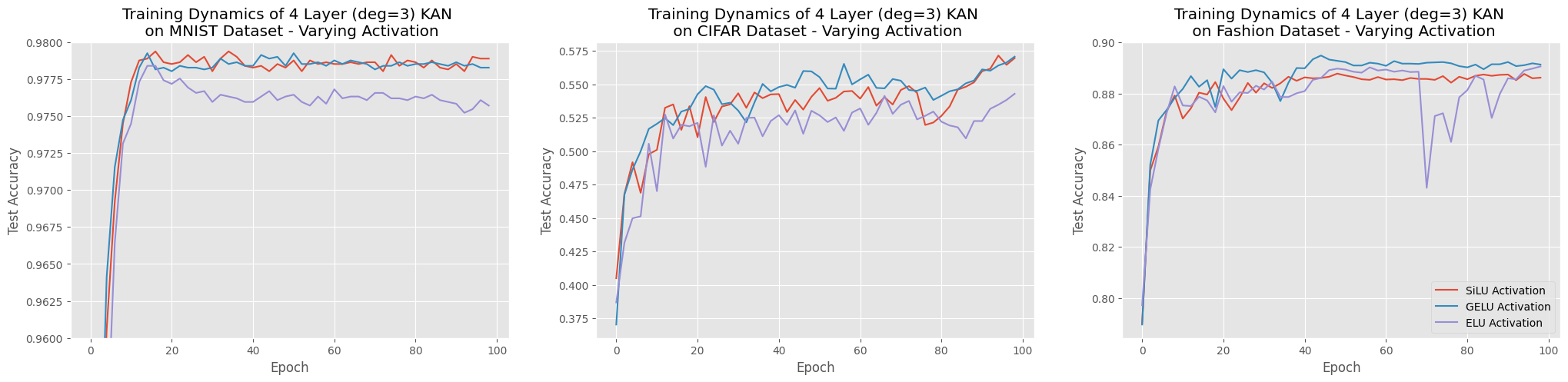}
\caption{The effect of activation function on KAN performance}
\centering
\end{figure}

Results indicate that KANs almost universally perform best when using the GeLU activation as compared to the SiLU or ELU. 

\newpage

\subsection{KANs - Scaling with Degree}
For a fixed architecture, we can also opt to add more interpolation points for each B-spline and increase the degree. As each spline fit becomes more exact, we expect test error to lower to a point and then increase as over-fitting starts to occur. It is tempting to view increasing the degree of KA units in a layer as analogous to increasing the width of the layer, however this hypothesis needs to be tested. We test a KAN with a single hidden layer and alternatively increase the hidden layer's width or the degree of the KA units. We fix the training scheme to the best one found in the above evaluations (Kaiming Normal initialization, AdAM optimizer, 1e-4 learning rate)

\begin{figure}[h]
\centering
\includegraphics[scale=0.30]{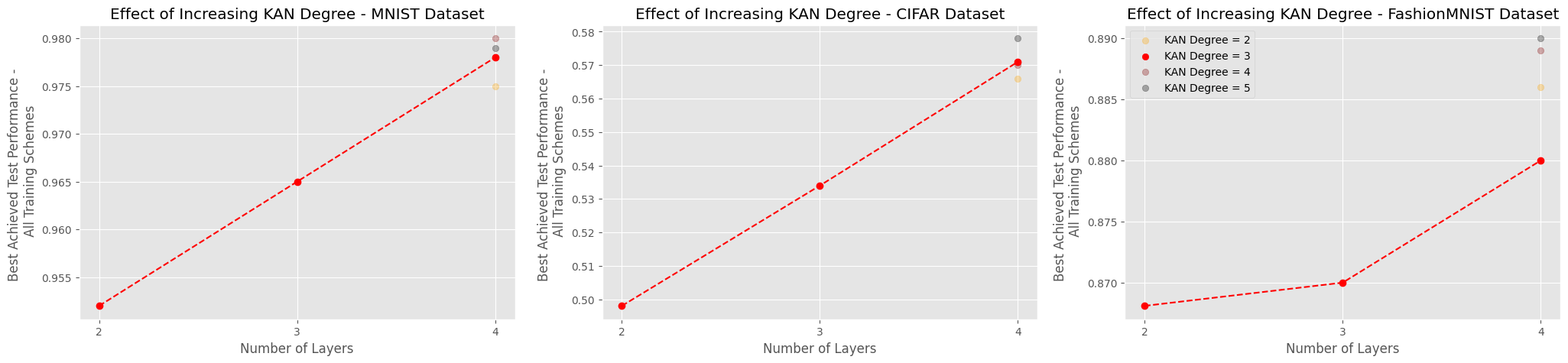}
\caption{The effect of B-spline degree and number of layers on KAN performance}
\centering
\end{figure}

The resulting test accuracies indicate that you cannot simply trade width for degree in KANs, and (on a per-parameter basis) increasing the degree in a layer seems to provide less performance utility than adding another layer of equal degree. This may not hold for deeper networks.

\section{Performance Observations}
\subsection{Efficiency}
We seek to compare model efficiency in a way that is hardware agnostic and dataset agnostic. It should capture intrinsic tradeoffs in model design and training schemes. To do this, we develop the following measure:

$EF(M, T, D) = \frac{A^{*}(M,T,D)}{E_{A*}(M,T,D) + 1}*\frac{1}{\log({ID(D) - P(M)}) + 1}$

Where $ID(D)$ is the intrinsic dimension \cite{int_dimension} of the test set $D$, $A^{*}(M,T,D)$ is the best test accuracy achieved by a model $M$ on test set $D$ under training scheme $T$, $E_{A*}(M,T,D)$ is the epochs taken to reach the best test accuracy, and $P$ is the number of parameters in the model. Note that this function is only defined when $P(M) \geq ID$, and is maximized when a model with $ID$ parameters reaches $100\%$ accuracy without training. This concept is beneficial in model selection for practical problems, for example while the best model for MNIST reaches $~99\%$ test accuracy, it may be more interesting to study the model with the highest efficiency (i.e the 25k parameter, single hidden layer MLP that reaches $95\%+$ accuracy with a single epoch of training). Note that this measure depends on the dataset only through the difference of the $ID$ term and the model size, capturing the fact that more complex datasets require more complex models to reach the same efficiency. 

We see that when comparing architectures with an equal number of parameters (mlpWide 2layers and kan 2layers), the efficiency of KANs is higher. However, there is also higher average generalization error (for the KAN) with an equivelent number of parameters. Inspection of individual training runs shows this to be due largely to KANs reaching their highest accuracies early in the training process, then overfitting to training data. We see that as we add layers, efficiency of KAN architectures seems to diminish and generalization error stays the same or increases, further reinforcing the notion that vanilla KANs may not scale as well to deep architectures compared to MLPs. 

In the graph below, model size is plotted against efficiency for all datasets jointly. $ID$ estimates for CIFAR and MNIST are as calculated in \cite{int_dimension}. 

\begin{figure}[h]
\centering
\includegraphics[scale=0.5]{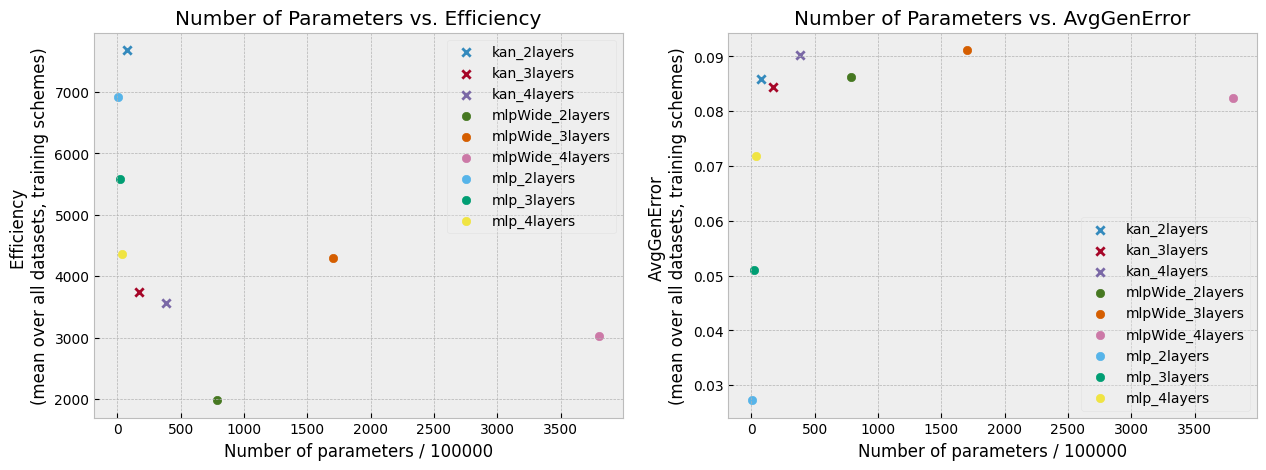}
\caption{Efficiency and Generalization Error trends across models and datasets}
\centering
\end{figure}

\newpage
\section{Conclusion and Future Work}
Our study demonstrates that KANs provide competitive accuracies when compared to MLP architectures, but are more sensitive to choice of training scheme. For optimal performance KAN architectures must be trained using adaptive optimizers and low learning rates. Without such training schemes, KANs tend to overfit and exhibit high generalization error (compared to MLPs trained with the same  scheme). Furthermore, KANs appear to be marginally more parameter efficient than MLPs, suggesting the possibility of smaller models to attain competitive accuracies. Because of a tendency to overfit much earlier during the training process than MLP architectures, caution must be used when scaling to very deep architectures using traditional MLP training schemes (future work evaluating residual connections in KAN architectures may help to alleviate some of these issues). These conclusions seem to hold regardless of training with backpropagation or the HSIC Bottleneck algorithm, and even when the number of parameters and number of layers is held constant. This suggests that while final accuracies remain competitive, KANs merit additional study on optimal training schemes to yield their full potential.  \newpage

%Bibliography

\bibliographystyle{unsrt}  
\bibliography{references}

\end{document}